\documentclass[conference]{IEEEtran}
\usepackage{epsfig}
\usepackage{graphicx}
\newcommand{\etal}{\textit{et al.}}

\usepackage{float}
\usepackage{adjustbox}
\usepackage{makecell}

\usepackage{array}
\newcommand{\minisection}[1]{\vspace{0.04in} \noindent {\bf #1}\ \ }

\ifCLASSINFOpdf
\else
\fi
\begin{document}
\title{R-PHOC: Segmentation-Free Word Spotting using CNN}

\author{\IEEEauthorblockN{Suman K. Ghosh}\IEEEauthorblockN{Ernest Valveny}
\IEEEauthorblockA{Computer Vision Center, Barcelona\\
\{sghosh,ernest\}@cvc.uab.es
}
}
\maketitle

\begin{abstract}
%Segmentation free word spotting in handwritten documents is a challenging task due to different factors like diverse handwriting styles, cursiveness, ageing of documents etc.
%In case of computer vision most tasks achieved great results using CNN
This paper proposes a region based convolutional neural network for segmentation-free word spotting. Our network takes as input an image and a set of word candidate bounding boxes and embeds all bounding boxes into an embedding space, where word spotting can be casted as a simple nearest neighbour search between the query representation and each of the candidate bounding boxes. We make use of PHOC embedding as it has previously achieved significant success in segmentation-based word spotting. Word candidates are generated using a simple procedure based on grouping connected components using some spatial constraints. %For all images in the dataset, we first generate a set of word candidate bounding boxes and then use our R-PHOC network to generate PHOC embeddings for all the bounding boxes using a single forward pass.
Experiments show that R-PHOC which operates on images directly can improve the current state-of-the-art in the standard GW dataset and performs as good as PHOCNET in some cases designed for segmentation based word spotting.

%In recent years most vision tasks are dominated by CNN or CNN based features,
%however only a few attempts has been made to use CNN in case of handwriting specially in word spotting. In segmentation based word spotting this is reltively easy as one can 
\end{abstract}


% \fi
\IEEEpeerreviewmaketitle

\section{Introduction}
%This demo file is intended to serve as a ``starter file''
%for IEEE conference papers produced under \LaTeX\ using
%IEEEtran.cls version 1.8b and later.
%I wish you the best of success.
% Generic things 
Word spotting is the task of searching for a given input word over a large collection of manuscripts. Indexing and browsing over large handwritten databases is an elusive goal in document analysis. The straightforward option for this includes using state-of-the-art OCR technologies to digitise the documents and then applying information retrieval techniques for information extraction.
However, in case of handwritten manuscripts this strategy does not work well as OCR available for printed documents are not directly applicable to these documents due to challenges like diversity of the handwriting style or the presence of noise and distortion in historical manuscripts. %This problem becomes even more challenging in multi-writer datasets.
 
 Thus, word spotting has been proposed as an alternative to OCR, as a form of content-based retrieval procedure, which results in a ranked list of word images that are similar to the query word. The query can be either an example image (Query-By-Example (QBE)) or a string containing the word to be searched (Query-By-String (QBS)).
 %Methods following QBE paradigm presents a huge disadvantage in practical applications \eg in order to spot a word the user needs to first locate/input an instance of such word. On the other hand QBS methods allow the user to type the keyword to search in a 
%much more natural way.

Initial approaches on word spotting followed a similar pipeline as OCR technologies, starting with binarization followed by structural/layout analysis and segmentation at word and/or character level. Example of this type of framework are the works of \cite{Vinci,Serrano}. The main drawbacks of these methods come from the dependence on the segmentation step, which can be very sensible to handwriting distortions. Other initial attempts on QBS based methods relied on the extraction of letter or glyph templates, either manually \cite{Konidaris,Leydier} or by means of some clustering scheme \cite{Marinai,Liang}. Then these character templates are put together in order to synthetically generate an example of the query
word. Although such methods proved to be effective and user friendly, their applicability is limited to scenarios where individual characters can be easily segmented. 
More generic solutions have been proposed in \cite{Fischer,Frinken}, where they learned models for individual characters and the relationship among them using either an HMM \cite{Fischer} or a NN \cite{Frinken}. These models are trained on the whole word or even on complete text lines without needing an explicit character segmentation. They are used to generate a word model from the query string that has to be compared with the whole database at query time. Therefore, computational time can rapidly increase with the size of the dataset. 

In this context it can be mentioned that example based methods are in a clear advantage as they can represent handwritten words holistically by compact numeric feature vectors. In this direction the work of Rusi\~{n}ol \etal \cite{Rusinol} proposes a representation of word images with a fixed-length descriptor based on the well known bag of visual words (BoW) framework. Comparison between the query and candidate image regions can be done by a simple cosine or Euclidean distance can be used, making a sliding window over the whole image feasible. In addition, Latent Semantic Indexing (LSI) is used to learn a latent space where the distance between word representations is more meaningful than in the original bag of words space. In \cite{almazan2014a} Almaz\'an \etal proposed to use a HOG based word representation in combination with an exemplar-SVM framework to learn a better representation of the query from a single example. Compression of the descriptors by means of product quantization permits a very efficient computation over a large dataset in combination with a sliding window-based
search. 
\begin{figure*}
    \centering
    \includegraphics{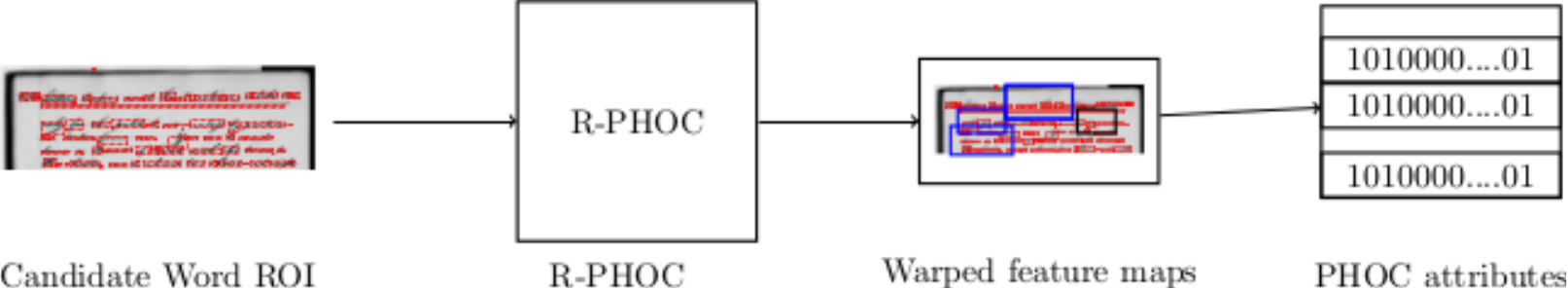}
    \caption{The proposed pipeline for segmentation free word spotting, First the whole document image is passed along with bounding box co-ordinates of candidate words, then the feature map corresponding to these regions are pooled by ROI pooling layer and finally passed to sigmoid activation to predict the PHOC attributes for the whole set of candidates. }
    \label{fig:schematic}
\end{figure*}
One of the challenges for QBS word spotting using compact word representations is to find a common representation that can be easily derived from both strings and images and permits a direct comparison between them. For that, Almaz\'an \etal\cite{almazan2014} proposed to learn a fixed length word representation based on character attributes to perform both QBE and QBS using the same framework. The attribute representation encodes the presence/absence of characters in different spatial positions in the word image through a Pyramidal Histogram of Characters (PHOC). Although originally the representation was learned using Fisher Vector as image features, some adaptations using CNNs to learn the attribute space have also been proposed \cite{PHOCNET,CNN-Ngram}. All the different variants of this representation have shown to be highly discriminant achieving state-of-the-art results in segmentation-based word spotting.

However using PHOC embedding in segmentation-free word spotting needs to embed all possible candidate words in PHOC space. In CNN based PHOC embedding \cite{PHOCNET,CNN-Ngram} this amounts to applying one CNN forward pass for each candidate (possibly in batch with GPU) and then computing the distance metric with the query. Fisher vector based PHOC was used for segmentation-free word spotting in \cite{ghosh2015sliding}.  Utilizing the additive nature of Fisher Vector embedding they propose to use a integral image of PHOC attributes to make the computation faster.
%can be done efficiently using integral image approach as used by \cite{}. Where PHOC embedding is computed on the whole page rather than individual words, which make the whole retrieval process faster. In this work our aim is to build such a system for retrieval which can take the whole image and produce PHOC embeddings for all candidate words in the document image.
%However, all these approaches also have a very high computational cost that makes their direct application to large datasets unfeasible in a segmentation-free scenario using sliding windows. 

In this work we propose a framework to extend in a more efficient way the PHOC word representation to segmentation-free word spotting leveraging more discriminative CNN features. To make the computations feasible, we take advantage of recent works in object detection \cite{R-CNN, fastR-cnn} that leverage a set of blind object proposals to find all instances of objects in an image using a single forward pass in a CNN. We take a similar approach for word detection using a set of candidate word proposals generated by grouping connected components based on a set of spatial constraints. The whole network (called R-PHOC) is trained end-to-end to generate the PHOC representation of every candidate region. Thus, given a query, word spotting can be performed by simply computing the distance between the PHOC representation of the query and the PHOC representation of all candidate regions obtained through the R-PHOC network. As these can be computed in a single forward pass in the network, the whole procedure is very efficient in terms of computation time. We evaluate our approach using standard George Washington showing state-of-the-art results for QBE word spotting.

The rest of the paper is organized as follows: In section \ref{sec:methodology}, we discuss the proposed methodology, which includes a sub-section \ref{sec:cand:gen}, \ref{sec:PHOC} \ref{sec:CNN-RPHOC} which respectively discuss the issues regarding generation of word candidates,  basic PHOC embedding and  details of training the PHOC embedding using region based CNN features. In section \ref{sec:exp} we show the results of the experiments carried out to validate our approach. Finally in section \ref{sec:conclusion} we report the conclusions of our work.
%is following section details the

% Object detections work a bit  introduce RCNN here

% Application of fast-rcnn here as object proposal are generated by using a simple connected components

% contributions

\section{Methodology}
\label{sec:methodology}
 An overall scheme of the framework is
illustrated in figure~\ref{fig:schematic}, the following sub-sections describes each componnents separately.
\subsection{Generation of candidate word regions}
\label{sec:cand:gen}
\begin{figure*}[ht]
    \centering
    \includegraphics[scale=0.19]{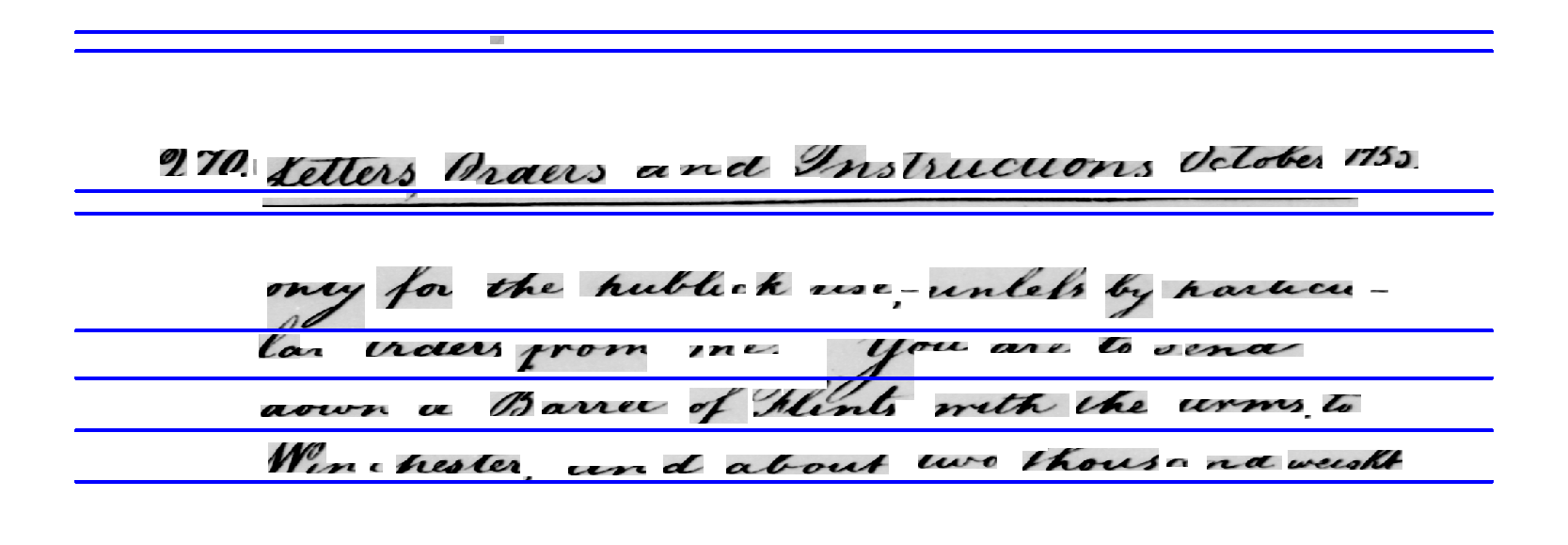}
    \includegraphics[scale =0.19]{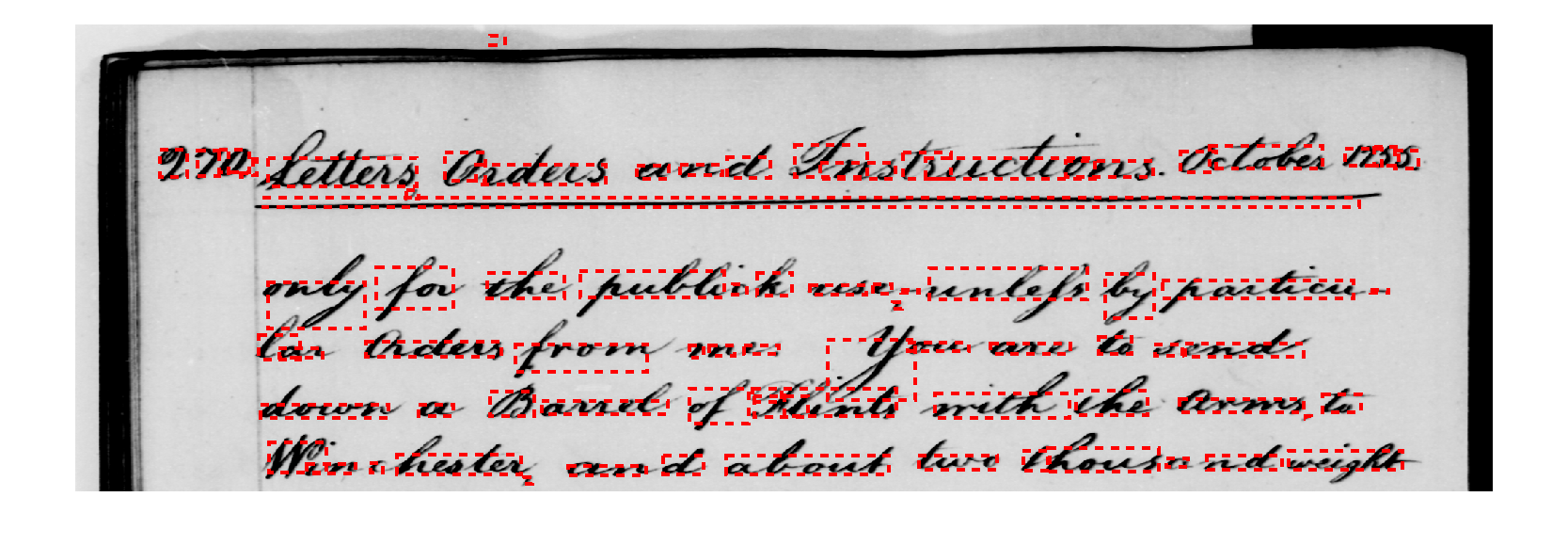}
    \caption{\textit{a}) Line estimation \textit{b}) Connected components for each line}
    \label{fig:Line_estimation}
\end{figure*}
 
In order to generate a set of blind candidate word regions over the whole image we rely on the analysis and grouping of connected components in such a way that we can guarantee a high recall in word localization. Connected components can easily be extracted from the document image. Moreover, in handwriting, connected components are mostly formed by pen strokes made by writers and thus atomic in nature. In general, they will not span more than a word. Therefore, any potential word in the document can be obtained by the combination of one or several neighbouring connected components. 

For connected component analysis, the document is binarized by setting the threshold as 75\% of the mean intensity of the image and then connected components are extracted using 8-pixel neighbourhood. As in this stage our goal is to retain as much information as we can in the images, rather than finding a clean binarized image, we use a high threshold to enhance the overall recall. Although this can lead to a larger number of false positives, the retrieval process can later discard them. 

For the combination of connected components into word candidate regions we will impose some spatial constraints on the total possible number of combinations to guarantee that candidate words are only composed of horizontally neighbouring connected components. 
The first constraint that we apply is co-linearity. For that, we avoid to use an explicit line segmentation method as usually, line segmentation methods are prone to errors and sensitive to noise. Our goal is not to find perfect line separation but rather to infer potential collinear connected components. 

In order to achieve that we first generate an over-complete set of line separation hypotheses by simply finding local minima in the horizontal projection of the image, after applying an average filter in order to smooth the projection profile . Then, every connected component will be assigned to all the lines for which they have a certain degree of overlapping. All connected components assigned to the same line will be considered as collinear. Let us note that one connected component can span more than one line hypothesis and therefore, can be combined with connected componentes in different lines. This is a way of assuring a high recall of word hypothesis.

In the process of finding minima of the projection profile, word ascenders and descencers can introduce some noise. Therefore, we pre-process connected components before computing the horizontal projection in order to remove this noise. For that, we replace every connected component by a minimal bounding box obtained following a greedy approach where we first compute the pixel density of the atomic bounding box. Then, starting from the middle of the bounding box we keep growing in both vertically and horizontally directions until we reach an area which contains 90\% of the pixel density of the corresponding bounding box.

Once each connected component has been assigned to one or more lines, candidate word regions will be generated as combinations of connected components within the same line. For that we will make use of the knowledge that words are arranged left to right in English and  will enforce some spatial ordering. Thus, we sort all connected components from left to right according to the $x$ position of their top left corner. Then, cnadidate word regions will be generated as any possible combination of consecutive connected components assigned to the same line. 

Finally, to reduce the computation time in further steps, we propose to use a simple and fast binary classifier in order to classify all these word candidate regions into word/non-word and filter non-word regions.  To learn this binary classifier we use simple features, which can be computed very fast. For each candidate region we compute a fixed length feature vector in the following way. Every candidate is divided vertically into $P$ segments and horizontally into $Q$ segments. For each segment the pixel density is calculated. This gives a $P + Q$ dimensional feature vector. We add to this feature vector the height and the width to the text box proposal normalize with respect to the average line height and width of all the text box proposals in the training set, respectively. This leads to a final feature vector of length $P + Q +2$. A linear support vector machine binary classifier is learned using these features.

\subsection{PHOC embedding}
\label{sec:PHOC}
PHOC representation\cite{almazan2014} provides an excellent embedding space where both strings and word images can be represented as low dimensional points. Once queries and candidate words are encoded in this embedded space word spotting is reduced to a nearest neighbour problem. 

The PHOC representation is based on the concatenation of spatial binary histograms of characters, encoding which characters appear in different positions of the string. The basic representation is just a histogram of characters over the whole string. In order to increase the discrimination power of the representation, new histograms are added at different levels in a pyramidal way to account for differences in the position of characters. Thus, at level 2, the word is split in two halves and the same histogram of characters is computed for each of the two halves. At level 3, the word is split in 3 sub-parts, at level 4 in 4, and so on. At the end, all histograms are concatenated in a single final word representation. In practice, 5 levels of decomposition are used and the histogram of the 50 most common English bigrams at level 2 is also added to the final representation to capture some relationship between adjacent characters, leading to a final word representation of $604$ dimensions.

For images, each dimension of the representation each dimension is an attribute encoding the probability of appearance of a given character in a particular region of the image, using the same pyramidal decomposition as in the PHOC representation. Originally each attribute was independently learned using an SVM classifier on a Fisher Vector description of the word image, enriched with the $x$ and $y$ coordinates and the scale of the SIFT descriptor. Recently the same embedding has been extended by Sebastian \etal in \cite {PHOCNET}, where they used CNN features in place of Fisher Vectors, improving overall accuracy. They trained a CNN to predict the estimated PHOC representation of a given input image, changing the usual softmax layer used for classification by a sigmoid activation function which is applied to every element of the output. They also used Spatial Pyramid Pooling to be able to feed images of different sizes to the network. In the next section we give more details about this network and how we have used it in our R-PHOC architecture.
%vector that correspond to one dimension of the PHOC representation.
% The network as whole is discussed later so may be not needed here
%The architecture of the PHOCNET network is based on a series standard $3\times3$ convolutional filters followed by Rectified
%Linear Units (ReLU) with an and an increasing number in the
%higher layers. In order to be able to feed images of different sizes to the netword the Spatial Pyramid Pooling proposed in \cite{SpatialPyramidPooling} is used. Training the CNN with PHOCs as
%labels can be seen as a multi-label classification task. Thus,
%the classical softmax function is substituted by a sigmoid activation function which is applied to every element of the output
%vector that correspond to one dimension of the PHOC representation. The network is trained using cross-entropy loss. After training, the PHOCNet outputs the estimated PHOC representation
%for a given input image. 

\begin{figure*}
    \centering
    \includegraphics[scale=0.14]{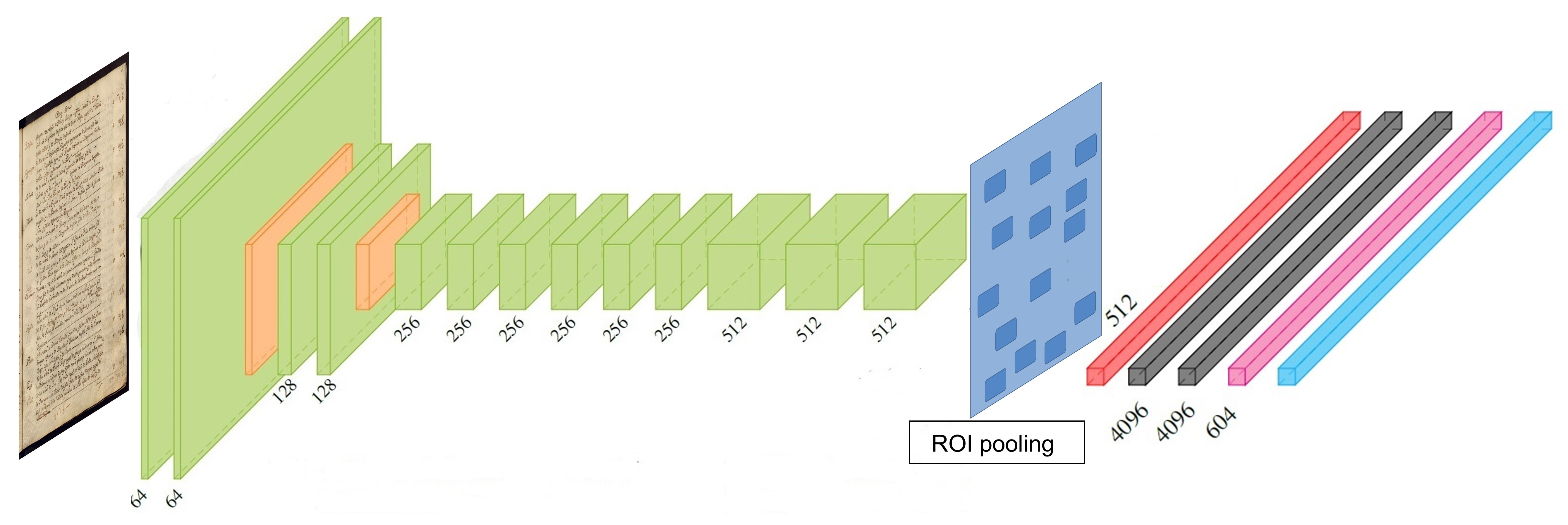}
    \caption{Architecture of our proposed R-PHOC network}
    \label{fig:network}
\end{figure*}
\minisection{Architecture of the R-PHOC network}

\subsection{PHOC embedding for regions}
\label{sec:CNN-RPHOC}

The task of word spotting is in a way similar to that of object detection, where the task is to find salient objects in an image and classify them into some predefined categories.  Traditionally object detection using CNN features was performed using a sliding window protocol over the input image. However like any sliding window based approach this involves lot of redundant computation.
To do this efficiently an specific architecture is needed, which can take as input the entire image and produce labels for all salient objects. The first breakthrough in this direction was made by Girschik \etal in \cite{R-CNN}, where they proposed the idea of region based CNN features and then SVM classifiers to classify regions into salient objects. A further modification of this approach was proposed in \cite{fastR-cnn}. In Fast-RCNN \cite{fastR-cnn} the concept of region of interest(ROI) pooling was first introduced, which enables the network to aggregate features from different salient regions of the image without needing to feed all the regions separately to the network. % and classify different regions of the same image
%which evolved to fast and faster R-CNN approaches.
Additionally the network is trained to regress the bounding box to predict more precise locations of salient objects. The whole network is trained end-to-end using a single multi-task loss function. In a nutshell a fast-RCNN framework takes a set of class independent object proposals as input and provides two set of outputs for each proposal: classification scores for each object category and an Offset for bounding box regression for each category of objects.

In this work we adopted this framework for handwritten text spotting. However, as the number of different English words is very large, text detection can not be seen as a classification problem.
To deal with this we train the network to predict the PHOC representation of the candidate window.
%rather than probability for a single category.  
Thus, this can be seen as an extension of the PHOC framework to a segmentation-free scenario using region based CNN features, thus the name R-PHOC is coined for our network. In the following sub-section the proposed architecture is described. % of our network.

The proposed architecture is given in Figure \ref{fig:network}. %We adopted the convolutional layers from the PHOCNET architecture proposed by Sebastian \etal in \cite{PHOCNET}. 
% are reused for a different problem.
 %nas we intend to use the pre-trained model, however to train the network in segmentation free scenario , the following changes are made :
%However as the goal of our work is to do a segmentation free word spotting, and we want to input the whole image as input to the network and want to predict PHOC for each candidate window using one forward pass of the network, the following changes are applied: % based on the candidate regions need
Thus architecture of the network contains the same set of convolutional and max pooling layers used \cite{PHOCNET} in order to to leverage their pre-trained models. This is done purposefully to be able to use transfer learning, i.e. to reuse the learned weights from their pre-trained network, which is a common practice in other computer vision tasks like image classification etc. However for simplicity and to be able to back-propagate the gradients the spatial pyramid layer is replaced by a ROI pooling layer in order to obtain the features for every candidate. Finally the feature representation produced by the ROI layer is fed to sigmoid activation layer.
% is applied to convert the feature space into PHOC space.  
%1) ROI pooling layer is introduced after the final convolutional layer.
%2) Sigmoid sit for segmentation free scenario on full images we   We adopted the convolutional layers from PHOCNET architecture. We use the same  $3 \times 3$ convolutional layers after a series of convolutional layers, to pool the region based features we use the ROI pooling and then for each pooled region a sigmoid layer is added. The architecture is shown in the figure \ref{}.

The ROI pooling layer uses max pooling to obtain a feature map from a ROI, by dividing the region of interest into sub-windows of fixed size. For example a ROI of $(x,y,h,w)$ is divided into grids of approximate size $h/H$ and $w/W$ and pooling the features from each grid into the corresponding output grid cell. Similar to standard max pooling, feature from each channel is pooled independently.
%Pooling is applied independently to each feature map channel as in standard max pooling. 

Our goal is to train the network to predict the PHOC embeddings for each ROI (word candidate region) using the CNN features. Then, the problem is different to most classification problems. Instead of having one one true class for every training example in PHOC there can be multiple positive classes (PHOC attributes) for every training sample. For classification problems a softmax layer is the de-facto standard to obtain the final output. However for multi-label classification tasks this can not be used. This is dealt in \cite{PHOCNET} by using a sigmoid activation function in place of the softmax layer. In this work we also make use of sigmoid activation functions to obtain the final output of the network.
%strategy to deal. 

Sigmoid cross entropy loss (or logistic loss) is used for training the network. Thus, the loss between a predicted PHOC $\hat \sigma$ and the real PHOC $\sigma$ of a given image is given as:

\begin{equation}
   l(\hat \sigma,\sigma) = -\frac{1}{n}\sum_{i=1}^{n}{\sigma_i\log(\hat\sigma_i) +(1-\sigma_i)(\log(1-\hat\sigma_i)}   , 
\end{equation}
where $n$ is the number of attributes in the PHOC representation $\sigma$. Given an image with a set of candidate words (ROIs), the loss is calculated as the summation of the individual loss for each ROI. Though the network could be used to regress the bounding box of every ROI to get a more precise location of every word, we leave that for future work.

\section{Training}
Training a deep network takes significant effort. Here, some of the hyper parameters are discussed in order to help the reproducibility of our method.
As the loss function adopted in this work is differentiable, the network can be trained end-to-end by back propagation. We used stochastic Gradient Descent with a learning rate initially fixed at 0.0001 and updated every 1000 iterations. We trained the network for 30000 iterations.
As the size of the document images in comparison to the images in Pascal dataset for object detection are much bigger, we divide the image into overlapping segments of size $600 \times 1000$ to be able to load the images in a GPU. One training epoch takes 0.524 seconds on average. We use a batch size of 128, i.e in one minibatch 128 ROIs are processed. The minibatches are sampled so as at least 60\% of them contain valid text regions with an overlap of more than 0.5. From the candidate word regions obtained using the procedure described in section \ref{sec:cand:gen}, we only used the candidates with more than 50\% overlap with ground truth words as text class and with less than 20\% overlap as background. The text candidates with text overlap between 20\% to 50\% are filtered as although they contain some text they do not constitute any valid word (e.g two consecutive words can be one candidate). Thus predicting PHOC for such candidates can act as a distractor. 

%The real PHOC's are obtained from the ground truth of the training samples(details are in the experimens section).

%from a spatial extent where the features are pooled in spatial  $H \times 7$ pooling 

%for segmentation free scenario, however we make some significant changes. 

%\begin{figure*}
%    \centering
%    \includegraphics[scale=0.13]{GW_page.eps}
%    \hspace{0.5mm}
%    \includegraphics[scale =0.17]{BCN.png}
%    \caption{Example pages from GW and BCN}
%    \label{fig:dataset}
%\end{figure*}

%whole document page is  

%disadvantage that in case of segmentation free scenario every possible candidate words need to be embedded in to PHOC space which is a time consuming process.  

\section{Experiments}
\label{sec:exp}

\subsection{Dataset}

\textbf{The George Washington (GW) dataset} \cite{Rath} has been used by most researchers in this field and has become one of the most important datasets to benchmark the results. This dataset contains 4860 words annotated at word level. The dataset comprises 20 handwritten letters written by George Washington and his associates in the 18th century. The writing styles present only small variations and it can be considered as a single-writer dataset. 

%The \textbf{Barcelona Centuries of marriage dataset (BCN)}\cite{BCN_database} contains 50 handwritten pages from a collection of marriage licences written in 1617 from Barcelona cathedral. This is a multi-writer database that contains 13000 words and have a bigger lexicon than GW. As the manuscripts are ancient the dataset also include a considerable amount of noise caused by ageing of manuscripts.

%Figure \ref{fig:dataset} shows one example page from each of the datasets.

\subsection{Experimental Protocol}

We performed Query-By-Example word spotting on the GW dataset following the standard protocols.  A retrieved candidate is considered positive if it overlaps with some ground truth word for which the labels are same and the Intersection over union between the ground truth word and the retrieved word region is more than some threshold. In our experiments we set this threshold to 50\% as it is usual practice in most cases. In GW, as the number of pages of the dataset is very small, we performed 4-fold cross validation, so that all words can be evaluated. We randomly divide the 20 pages into 4 bins of 5 pages each, and therefore for each fold, we used 15 pages for training and 5 for test. While testing we only used words from the test pages and the list of queries was also formed by the words from these 5 pages. On average every fold contains 1230 words. 

\subsection{Results}

For baseline analysis we perform QBE word spotting in GW dataset for segmented words, i.e each segmented word is assumed as a ROI and every document page is passed through the network once. The results for this experiment are shown in \ref{tab:baseline}. Though the goal of this work is not segmentation-based word spotting, we did this study to analyse the effect of region based CNN features. As we can see that in segmentation-based scenario the R-PHOC method reaches 92.75\% mAP which is comparable to FV-PHOC of \cite{almazan2014}. R-PHOC performs slightly worse than PHOCNET, which is expected as in region based approaches features are integrated in discrete intervals. However, as described above this is very efficient in comparison to other PHOC based approaches as in one forward pass all candidate words of the entire document can be evaluated, thus achieving a massive parallelism.

%method this method with oth

\begin{table}[h]
\label{tab:baseline}
\centering
\begin{tabular}{l|c}
\hline
\textbf{Methods} & \textbf{Mean Average Precision} \\%& \textbf{SVT-Full} & \textbf{SVT-50} \\
 \hline
PHOCNET \cite{PHOCNET}   &96.71 \\%& 71.3 &87.4    \\
FV-PHOC \cite{almazan2014}  &93.04 \\%&82.38&91.44\\
R-PHOC (Region based CNN) &92.75\\ %&81.14&89.79\\
%Proposed (LSTM + attention model+LM)&33.67\\ %&81.35&90.04\\
%Proposed (LSTM + attention model+LM+FT)&43.86\\
%No Filtering&1025&80.65&56.72&66.60\\
\hline
\end{tabular}
\caption{Performance of R-PHOC for QBE word spotting in segmented word images)}
\end{table}

\minisection{QBE word spotting in segmentation-free scenario}
Table \ref{table:comparisonQBE} summarises the results of applying the R-PHOC network to segmentation-fre QBE word spotting, compared to other methos in the state-of-the-art.

For a better comparison of the effect of using region based features we also provided the results of performing segmentation-free QBE word spotting using the original PHOCNET model. To generate these results we used the same candidates as in our approach, but in this case the PHOC representation is obtained by processing each candidate one by one. As our network is also learned to predict PHOC given a ROI of an image, the baseline result of PHOCNET is a way to provide an upper limit to our network.  However this comes with a huge cost of applying the CNN for every candidate, while our network obtains the PHOC representation of all candidates in a single forward pass.

This can be verified in Table \ref{table:comparisonQBE}. First of all, we can see that all results using PHOC as the basis for word representation clearly outperform the rest of methods. The baseline consisting of applying PHOCNET separately to all word candidate regions achieves an accuracy of 87.71, but the cost of applying the CNN to all individual candidate makes this approach unfeasible in terms of computation time. When applying R-PHOC the performance decreases to 79.83, still outperforming all other methods by a significant margin, but reducing by a factor of almost 10  the computation time required. This result shows that the using ROI pooling permits to obtain a high accuracy while enabling the computation of the signature from all word candidates from an image by a single forward pass, which makes this approach computationally efficient. We have noticed that the down-sampling in ROI pooling has a larger effect on the small words. A similar phenomenon is observed in the case of object detection where small objects remain undetected by state-of-the-art CNN detector. Thus, we also evaluated our network on words which contain more than 5 letters and obtained an accuracy of 86.7, which is comparable to the baseline PHOCNET. 

%As one of the main goal of this work is efficiency with respect to retrieval time, we compare the query evaluation time in table \ref{table:comparisonQBE} we also compared the average query evaluation time with PHOCNET. As expected a query in R-PHOC can be evaluated much faster than that of PHOCNET.  

\begin{table*}[ht]
%\rotatebox{90}{
\begin{adjustbox}{width=\textwidth,center=\textwidth}
\centering
\scriptsize
\begin{tabular}{c|c|c|c|c}
\hline
 Method & Segmentation&Dataset&Accuracy& Query evaluation time (in sec)\\ %&Org. Comp. Time.&Proposed Time \\
 \hline
 Recurrent Neural Networks \cite{Frinken} & Line Level&20 pages, all
words of the
training   set   appearing  in  all 4
folds as queries &71\% mean prec. & \\
 Character HMMs \cite{Fischer} & Line Level&all 
words     of     the
training   set   ap-
pearing  in  all  4
folds as queries&62\% mean prec.&\\
 Slit style HOG features \cite{slitstyle}& Line Level &20 pages, 15 queries&79.1\% mAP\\
 Gradient features with elastic distance\cite{Leydier2}&NO&20 Pages, 15 Queries&60\%\\
 BOVW+HMM \cite{Rothacker} & No&5 pages Test and 20\% overlap is considered as true positive &61.35\% mAP\\
Exemplar SVM \cite{almazan2012bmvc}& NO& All ground truth words as queries&54.5\%\\ 
bag-of-visual-words+LSI \cite{Rusinol}&NO& All Ground Truth Words as Queries&61.35\% mAP\\ \hline
Baseline PHOCNET&NO&All Ground Truth Words as Queries&87.71\%mAP&371 sec.\\
R-PHOC (Proposed)&NO&All Ground Truth Words from test set as queries&79.83\%mAP&48sec.\\
R-PHOC (Proposed) $>5$ characters&NO&All Ground Truth Words from test set as queries&86.7\%mAP&48sec.\\
%BOVW-HMM \cite{Rothacker}&61.1\% mAP&83.7\%&\\
\hline
\end{tabular}
\end{adjustbox}
%}
\caption{Comparison with state of the art methods for QBE in GW20}
\label{table:comparisonQBE}
\end{table*}

%A

\section{Conclusions}
\label{sec:conclusion}
An efficient CNN based segmentation-free word spotting is proposed.
We apply a simple pre-segmentation to generate a set of word candidates which are then passed through a convolutional neural network to predict PHOC embedding for all candidates in a single forward pass in the network. Word spotting is then performed using nearest neighbour approach.
We observed that region based features obtained by integrating CNN feature maps can be used to train PHOC embeddings, thus generating an efficient scheme for segmentation-free word spotting.
Though the proposed approach performs better than most of the approaches using hand crafted features, our method still performs very poorly in case of words with a low number of characters. In the future this needs to be taken care of by applying feature maps of various sizes.
Another future direction could be automatically generate candidate bounding boxes by means of regression using the same architecture.

\end{document}